\documentclass[12pt]{article}

\usepackage{sbc-template}

\usepackage{graphicx,url}

\usepackage[utf8]{inputenc}  
\usepackage{amsmath,amssymb,amsfonts}
\newcommand{\bftab}{\fontseries{b}\selectfont}

\usepackage{algorithm}
\usepackage{algpseudocode}

\title{
A Network-Based High-Level Data Classification Algorithm Using Betweenness Centrality 
}

\author{Esteban Wilfredo Vilca Zuñiga\inst{1}, Liang Zhao\inst{1} }

\address{
Dept. of Computing and Mathematics\\
FFCLRP-USP\\
Ribeirão Preto, Brasil\\
\email{ evilcazu@usp.br, zhao@usp.br}
}

\begin{document} 

\maketitle
\begin{abstract}
Data classification is a major machine learning paradigm, which has been widely applied to solve a large number of real-world problems. Traditional data classification techniques consider only physical features (e.g., distance, similarity, or distribution) of the input data. For this reason, those are called \textit{low-level} classification. On the other hand, the human (animal) brain performs both low and high orders of learning and it has a facility in identifying patterns according to the semantic meaning of the input data. Data classification that considers not only physical attributes but also the pattern formation is referred to as \textit{high-level} classification. Several high-level classification techniques have been developed, which make use of complex networks to characterize data patterns and have obtained promising results.
In this paper, we propose a pure network-based high-level classification technique that uses the betweenness centrality measure.
We test this model in nine different real datasets and compare it with other nine traditional and well-known classification models. The results show us a competent classification performance. 
\end{abstract}

\section{Introduction}
Machine learning can be defined as a set of methods that can automatically detect patterns in data, and then use the uncovered patterns to predict the future data, or perform other kinds of decision making under uncertainty \cite{murphy2013machine}.

Usually, machine learning is divided into three main paradigms: \textit{supervised learning}, \textit{unsupervised learning}, and \textit{semi-supervised learning} \cite{geron2017homl}. Supervised learning uses tagged data to detect patterns and predict future cases. According to the kind of labels, the prediction is called \textit{classification} for categorical labels and \textit{regression} for numerical labels.
The unsupervised machine learning algorithms explore the data to search for possible structures that can tag the data. For example, on social media, the users don't provide necessarily some special information like political preferences. However, using the collected data an unsupervised algorithm could detect this \cite{geron2017homl}. Semi-supervised learning is a combination of supervised and unsupervised learning. Usually, the quantity of labeled data is low because tagging is expensive. So, semi-supervised algorithms use some labeled data to predict the untagged data \cite{ankur2019hulup}.

Data classification is one of the most important topics in supervised learning. It aims at generating a map from the input data to the corresponding desired output, for a given training set. The constructed map, called a classifier, is used to predict new input instances. 

Many algorithms use only physical features (e.g., distance, similarity, or distribution) for classification. These are called \textit{low-level} classification algorithms. Such algorithms can get good classification results if the training and testing data are well-behaved, for example, the data satisfies a normal distribution. But, these techniques have a problem to classify data with complex structures. On the other hand, a high-level algorithm uses data interaction as a system for classification, exploiting the structure of the data to capture patterns. In this way, it can perform classification according to pattern formation of the data like cycles, high links density, assortativity, network communication (\textit{betweenness}), and so on instead of just measuring physical features like euclidean distance.  

In order to capture the structure and properties of the data, we propose to work with complex networks, which are defined as large scale graphs with nontrivial connection patterns \cite{barabasi2002smcn}. Many network measures have been developed and each one of them characterizes network structure from a particular viewpoint. 
In the category of degree-related measures, we have the \textit{density} that represents how strong the nodes connections are \cite{zhao2016mlcn} and \textit{assortativity} degree that represents the attraction of nodes with a similar degree (degree correlation) \cite{Newman2003assort}. 
In the category of \textit{centrality measures}, we have \textit{betweenness centrality} that measures the node importance for communication on the network \cite{freeman77}, \textit{closeness vitality} that measures the impact of a network communication if a node is removed \cite{zhao2016mlcn}, and so on.

Several high-level algorithms have been proposed to use network measures to make a classification. \textit{Impact measure} approach tries to reduce the variation of a measure once a new node is inserted into a network \cite{tiago2018hldc}, \textit{link prediction} approach uses a meta class node to represent each label and the classification is performed using link predictions techniques \cite{seyed2019}, and \textit{importance measure} exploits the page-rank algorithm for classification \cite{carneiro2018importanceconcept}.

The technique proposed in this work, captures the structure of data using just one metric $betweenness\ centrality$. This measure captures the node importance for the graph communication. Nodes that have low $betweenness$ tend to be on the periphery on the contrary the nodes tend to be focal points \cite{zhao2016mlcn}. Instead of focusing on the node insertion impact or preservation of the structure measure using many network measures. We focus on the structure generated once a new node is inserted and identify if this inserted node presents similar features to the others in the new network. Also, unlike other methodologies that require a classical algorithm like SVM (Support Vector Machine) to complete the high-level classification \cite{thiago2012hldc}, our methodology uses pure network measures to classify. This approach shows good performance, avoids the double calculations of \textit{impact measure} method, reduces the number of properties to be used, and do not require to be combined with other classical techniques.

\section{Model Description}

In this section, we describe the working mechanism of our model. Firstly, we give an overview of the training and classification model phase. Then we provide details about each step of the algorithm. Finally, we describe how we use the betweenness measure on the model.

\subsection{Overview of the Model}
Each complex network consists of a set of nodes or vertices $\mathcal{V}$ and a set of links or edges $\mathcal{E}$ between each pair of nodes.
The input data $\mathcal{D}$ of $N$ elements for $supervised\ learning$ contains two parts: the attributes $\mathcal{X}$ and the labels $\mathcal{Y}$.

In $supervised\ learning$, the dataset $\mathcal{D} = \{ (X_1,y_1),...,(X_n,y_n) \}$ 
where $X_i=(x_1,...,x_d)$ represents the $d$ attributes, and $y_i$ represents the label of the instance $X_i$. The values of $y_i \in \mathcal{L}=\{ l_1,...,l_c \}$ where $\mathcal{L}$ is the possible labels of the instance. 
The goal of $supervised\ learning$ is to predict the $y_i$ values using the instances $X_i$. This could be considered as function approximation $f(X_i)\approx y_i$ where the function $f$ is our algorithm. To evaluate the model, it is required to split the data in training and testing datasets. The $X_{training}$ dataset will be used to build our model and the $X_{testing}$ dataset will be used for evaluation. 

In the training phase, we will build complex networks using the training dataset. The instances in the dataset will be the nodes and the links will represent the similarity between these nodes. Therefore, we will have $\mathcal{D} \mapsto \mathcal{G}=\langle \mathcal{V},\mathcal{E} \rangle$, where $\mathcal{V}=\{1,...,N\}$ is the set of nodes and $\mathcal{E}$ is the set of links in the complex network $\mathcal{G}$.
The links could be created using $kNN$ and $\epsilon-radius$ or personalized relation metrics like friendship on social data, flight routes, or city connections.

The network $\mathcal{G}$ will be built using $X_{training}$ to produce the nodes $\mathcal{V}$  and $kNN$ and $e-radius$ as relation metric for links $\mathcal{E}$. Then, we remove the links between nodes with different labels $y_i$. Following this strategy, we will have one network component  $\mathcal{G}^i$ for each label in $\mathcal{L}$.

In the testing phase, we insert a node from $X_{testing}$ into each component $\mathcal{G}^i$ following the same $kNN$ and $e-radius$ rules of training phase. Then, we calculate the $betweenness\ measure$ of this node in each $\mathcal{G}^i$.
This measure is compared to the others from each network component $\mathcal{G}^i$. So, the differences are saved in a new list for each $\mathcal{G}^i$. 

Finally, we get the average of the $b$ lowest values for each list and we classify the new node to the
$\mathcal{G}^i$ with the lowest average. Then, we remove this node from the other components.
In the case that the average differences of two or more lists are equal, we use the number of links connected to this new node in each component as a second difference measure. 

\subsection{Network-Based High Level Classification Algorithm Using Betweenness Centrality (NBHL-BC)}

The proposed high-level classification algorithm, which will be referred as NBHL-BC, has four parameters $k$, $e$, $b$, and $\alpha$. Where $k$ is the number of neighbors used in the $kNN$, $e$ is the percentile into $kNN_{distances}$ used to calculate $\epsilon$, $b$ is the number of nodes with similar $betweenness$ used for classification,and $\alpha$ is the weight to balance between links and $betweenness\ centrality$.

During the training phase, we need to build the network using the $X_{training}$ and $Y_{training}$ where $X_{training} \mapsto \mathcal{G} = \langle \mathcal{V,E} \rangle$. 
Each node in $\mathcal{V}$ is related with one instance in $X_{training}$ and each link in $\mathcal{E}$ is defined following these two techniques:

\begin{equation} \label{network_construction_rule}
  \mathcal{N}(X_i)=\begin{cases}
    \epsilon\text{-}radius(X_i,y_i), & \text{if }|\epsilon\text{-}radius(X_i,y_i)|>k\\
    kNN(X_i,y_i), & \text{otherwise}
  \end{cases}
\end{equation}

Where $(X_i,y_i)$ represents a pair of data instance $X_i$ and its corresponding label $y_i$.
For each instance $X_i$, $\mathcal{N}(X_i)$ is the set of nodes to be connected to it, its neighborhood. 
$\epsilon\text{-}radius(X_i,y_i)$ returns the set of nodes $\{ X_j, j \in \mathcal{V} : distance(X_i,X_j)< \epsilon \land y_i = y_j  \}$
i.e. the set of nodes $X_j$ whose similarity with $X_i$ is beyond a predefined value $\epsilon$ and have the same class label$\ y_i$. 
Here, $distance$ is a similarity function like euclidean distance. $kNN(X_i, y_i)$ returns the set containing the $k$ nearest neighbors of $X_i$. 
The value $\epsilon$ is the percentile $e$ of the $kNN_{distances}$ in the sub graph of $y_i$.
Note that the $\epsilon$-radius criteria is used for dense regions ($|\epsilon-radius(Xi)| > k$), while the $kNN$ is employed for sparse regions. 
With this mechanism, it is expected that each label will have an independent sub graph $\mathcal{G}^c$ \cite{thiago2012hldc} \cite{thiago2015turistwalk} \cite{tiago2018hldc}.

\begin{figure}[ht]
    \centering
    \begin{subfigure}[b]{0.4\linewidth}
      \centering
      \includegraphics[width=1\linewidth]{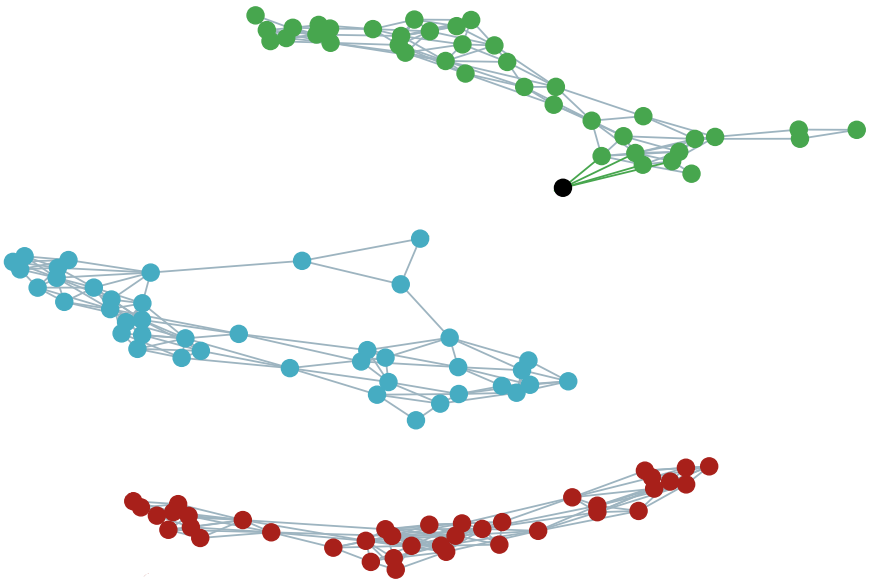}
      \caption{Inserted node in $\mathcal{G}^g$}
      \label{fig:insertionG}
    \end{subfigure}
    \begin{subfigure}[b]{0.4\linewidth}
      \centering
      \includegraphics[width=1\linewidth]{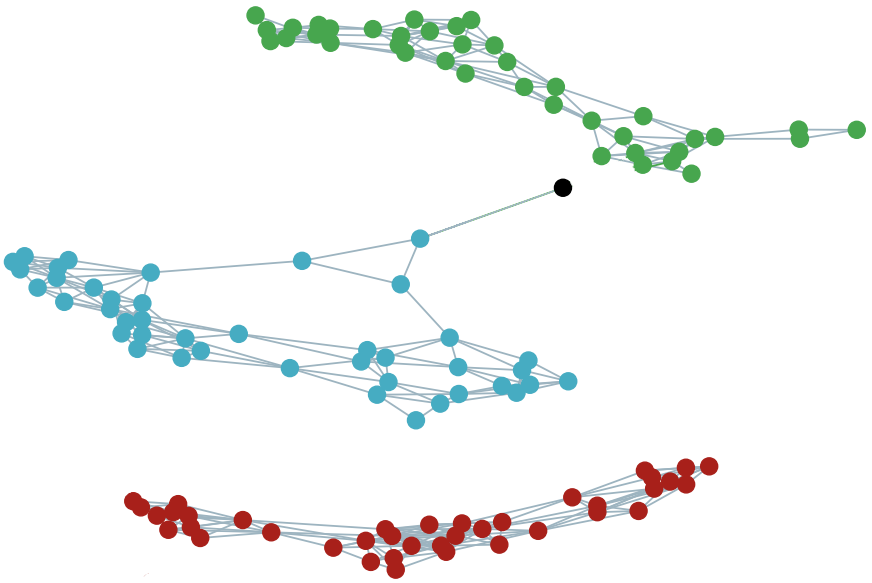}
      \caption{Inserted node in $\mathcal{G}^b$}
      \label{fig:insertionB}
    \end{subfigure}
    \begin{subfigure}[b]{0.4\textwidth}
      \centering
      \includegraphics[width=1\linewidth]{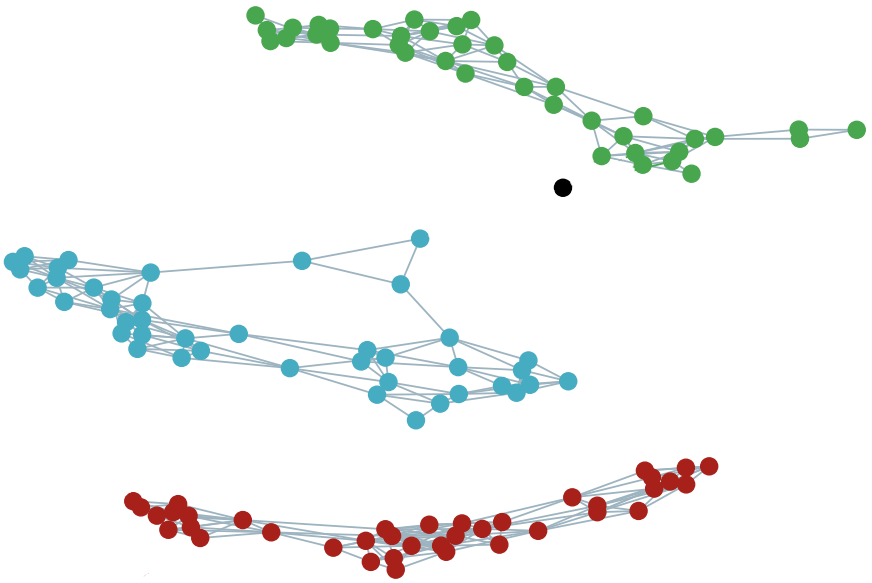}
      \caption{Inserted node in $\mathcal{G}^r$}
      \label{fig:insertionR}
    \end{subfigure}%
    \begin{subfigure}[b]{0.4\textwidth}
      \centering
      \includegraphics[width=1\linewidth]{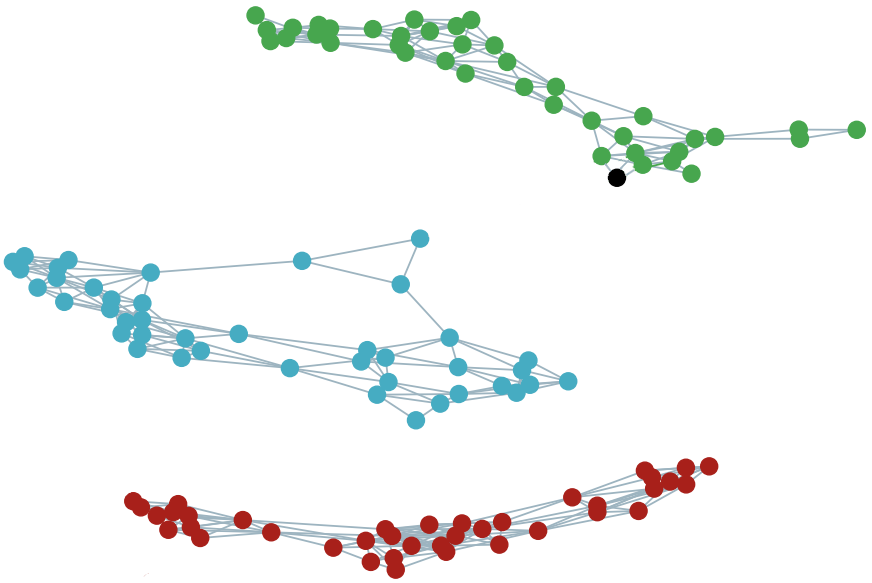}
      \caption{Final Classification}
      \label{fig:insertionFinal}
    \end{subfigure}
    \caption{Classification of a new instance (dark node) into the iris dataset graph $\mathcal{G}$ using with $k=5\ e=0.5\ b=5\ \alpha=1.0$ }
    \label{fig:classificationProcess}
\end{figure}
In Figure \ref{fig:insertionFinal}, we can see the graph  $\mathcal{G}=\{\mathcal{G}^r,\mathcal{G}^g,\mathcal{G}^b\}$ where $r,g,b$ represent the sub graphs with nodes red, green and blue. On the testing phase, we insert each $X_{testing}$ instance (dark node) to each component $\mathcal{G}^i$ following the same rule on equation (\ref{network_construction_rule}), and assuming that the node will be inserted in each sub graph. 

For example, in Figure \ref{fig:classificationProcess}, there are three network components $\mathcal{G}^i$ and the node to be tested is inserted to each one. In Figures \ref{fig:insertionG} \ref{fig:insertionB} \ref{fig:insertionR}, the node uses its $k$ nearest neighbors with the same label. In this case with $k=5$, there are 4 nodes in $\mathcal{G}^g$, 1 in $\mathcal{G}^b$, and 0 in $\mathcal{G}^r$. The  $\epsilon-radius$ with the $e=0.5$ (median of $kNN_{distances}$) is less than 5, because the current inserted node presents a sparse behavior; for this reason, we will use just $kNN$.Moreover, due to the condition of the same label, the algorithm will produce one sub graphs for each possible label.



    

Now we calculate the $betweenness\ centrality$ for the node in each component when the new node is inserted. Following this rule, the inserted node will have different values for each component. 

The $betweenness\ centrality$ is a mixed measure (global and local) that captures how much a given node is in the shortest paths of others nodes \cite{zhao2016mlcn}. This measure captures the influence of a node in the communication of the network \cite{needham2019graph}. We capture not only the characteristics of a node but also the behavior of their neighborhood. So, we have a metric that provides local and global network characteristics. This metrics is defined in the equation \ref{betweenness}.
\begin{equation} \label{betweenness}
    B(i) = \sum_{s\neq i\in\mathcal{V}} {\sum_{t\neq i\in\mathcal{V}}{ \frac{ \eta^{i}_{st}}{ \eta_{st}}}}
\end{equation}
where $\eta^{i}_{st}$ is 1 when the node $i$ is part of the geodesic path from $s$ to $t$ and 0 otherwise. $\eta_{st}$ is the total number of shortest paths between $s$ and $t$.

Then, we calculate the difference of this measure between the inserted node and the other nodes in each component $\mathcal{G}^i$. In the algorithm \ref{alg:Classification} on line 14, we can show this step. In the algorithm \ref{alg:Insertion}, we can appreciate how an inserted node will present a different $betweenness\ centrality$ for each sub graph.

\begin{algorithm}
\caption{Node Insertion}

\label{alg:Insertion}
\begin{algorithmic}[1]
\Function{NodeInsertion}{$\mathcal{G},instance,index,k,e$}
    \State $\mathcal{\langle V,E \rangle \leftarrow G}$ \Comment{ index is the number of nodes in the graph +1}
    \State $\mathcal{V} \leftarrow  \mathcal{V} \cup \{ index \}$ 
    \State $edges \leftarrow \{  \}$
    
    \For {$ X_i \in X_{training}$}
        \If{$ X_i \in \mathcal{N}(instance,k,e) $}
        \State $edges \leftarrow edges \cup (i,index)  $ 
        \EndIf
    \EndFor
    \State $\mathcal{E} \leftarrow  \mathcal{E} \cup \{ edges \}$ 
    \State $\mathcal{ G \leftarrow \langle V,E \rangle}$
    \State \Return $\mathcal{G}$ 
\EndFunction

\end{algorithmic}
\end{algorithm}

These values will be inserted into an independent list for each component $\mathcal{G}^i$. 
We will calculate the average of the $b$ lower values on each list. In the \ref{alg:Classification} on line 19, we can appreciate how we get just the b lower elements on $NB$ previously sorted on line 16. 
The results are stored on the array $\mathcal{W}=\{w_1,...,w_c\}$ where each $w_i$ represents the average difference of the $b$ nearest betweenness node values on the sub graph $\mathcal{G}^i$. This process is represented in the algorithm \ref{alg:Classification} on line 27 and 28.

\begin{equation} \label{normalizeW}
    \mathcal{W}^n=\frac{1-\mathcal{W}}{\sum_{w_i \in \mathcal{W}}{1-w_i}} 
\end{equation}
Where $\mathcal{W}^n$ is the normalized version of $\mathcal{W}$. In order to avoid conflicts of probabilities with the same value $w_i \in \mathcal{W}^n$, we calculate the number of links of the inserted node with respect to each sub graph $\mathcal{G}^i$ on the array $\mathcal{T}$.
Then, we follow a similar process of equation \ref{normalizeW} for $\mathcal{T}$ normalization.
This process is represented in the algorithm \ref{alg:Classification} on line 29.
\begin{equation} \label{normalizeT}
    \mathcal{T}^n=\frac{\mathcal{T}}{\sum_{t_i \in \mathcal{T}}{t_i}} 
\end{equation}
Finally, once we normalize these values, we calculate the sum of $\mathcal{T}^n,\mathcal{W}^n$ and made a final normalization.
\begin{equation} \label{normalizeH}
    \mathcal{H}=\frac{(\alpha)\mathcal{W}^n+(1-\alpha)\mathcal{T}^n}{\sum_{t_i \in \mathcal{T}^n,w_i \in \mathcal{W}^n}{(\alpha)w_i+(1-\alpha)t_i}} 
\end{equation}
where $h_i \in \mathcal{H}$ represents the probability of a node $i$ to be inserted in the sub graph $\mathcal{G}^i$, and $\alpha$ controls the weights between structural information and number of links. If $\alpha = 1.0$, we just capture information using $betweenness\ centrality$, and if $\alpha = 0.0$, we just capture information about number of links. The fully algorithm is described in algorithm \ref{alg:Classification}.

\begin{algorithm}
\caption{Classification Algorithm}
\label{alg:Classification}
\begin{algorithmic}[1]

\Function{Classification}{$\mathcal{G}$,$instance,k,e,b,\alpha$}
    \State $index \leftarrow n+1$ \Comment{ $n$ is the number of nodes in $\mathcal{G}$ }
    \State $ \mathcal{G} \leftarrow$NodeInsertion($\mathcal{G},instance,index,k,e$)
    \State $\mathcal{W} \leftarrow \{\}$
    \State $\mathcal{T} \leftarrow \{\}$
    \For { $\mathcal{G}^i \in \mathcal{G}$} \Comment{Where each $\mathcal{G}^i$ is a subgraph}
        \State $NB \leftarrow \{\}$ \Comment{NB is a list of node betweenness differences}
        \State $\mathcal{\langle V}^{i},E^{i} \rangle \leftarrow \mathcal{G}^i$
        \State $Links \leftarrow 0$
        \For{ $j \in \mathcal{V}^i $  }
            \If{ $j \in \mathcal{N}(index,k,e) $}
                \State $Links \leftarrow Links +1$
            \EndIf
            \State $NB \leftarrow NB \cup \{B(index)-B(j)\}$ \Comment{B is betweenness centrality}
        \EndFor
        \State Sort(NB) \Comment{NB has the differences between the nodes in $\mathcal{G}^i$ and the new node}
        \State $Total \leftarrow 0$
        \State $count \leftarrow 0$ 
        \While{ $ count < b$  }
            \State $Total \leftarrow Total + NB[count]$
            \State $count \leftarrow count + 1$
        \EndWhile
        \State $Total \leftarrow \frac{Total}{b}$
        \State $\mathcal{W} \leftarrow \mathcal{W} \cup Total$
        \State $\mathcal{T} \leftarrow \mathcal{T} \cup Links$
    \EndFor
    \State $\mathcal{W}^n \leftarrow 1-\mathcal{W}$
    \State $\mathcal{W}^n \leftarrow \frac{\mathcal{W}^n}{sum(\mathcal{W}^n)} $
    \State $\mathcal{T}^n \leftarrow \frac{\mathcal{T}}{sum(\mathcal{T})} $
    \State $\mathcal{H} \leftarrow (\alpha)\mathcal{W}+ (1-\alpha)\mathcal{T}$
    \State $\mathcal{H} \leftarrow \frac{\mathcal{H}}{sum(\mathcal{H})} $
    \State \Return MaxIndexValue($\mathcal{H}$) \Comment{$\mathcal{H}$ has each class probability}
\EndFunction

\end{algorithmic}
\end{algorithm}

\section{Performance Tests on Toy Datasets}

In this section, we present the classification performance of our algorithm in toy datasets and compare the results with other algorithms using python as programming language and Scikit-learn library for algorithms \cite{scikit-learn}. Specifically, we test our algorithm against Multi Layer Perceptron (MLP) \cite{RPROP-MLP}, Decision Tree C4.5 (DT) \cite{DecisionTree}, and Random Forest (RF) \cite{breiman2001random}.
The algorithms are tested using cross validation 10-folds, executed 10 times, and we use a grid search to select the hyper parameters that give the best accuracy for all the algorithms.

\begin{figure}[h]
\centering
\begin{subfigure}{0.4\textwidth}
  \includegraphics[width=\linewidth]{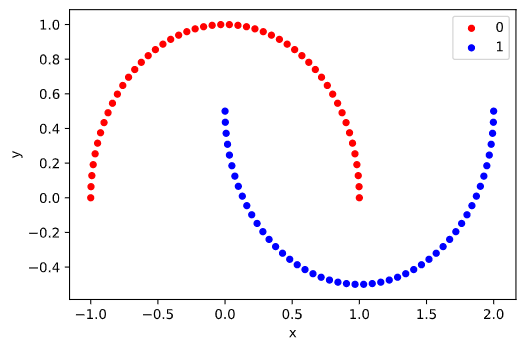}
  \caption{Moons Noise 0.0}
  \label{fig:moons0}
\end{subfigure}%
\begin{subfigure}{0.4\textwidth}
  \includegraphics[width=\linewidth]{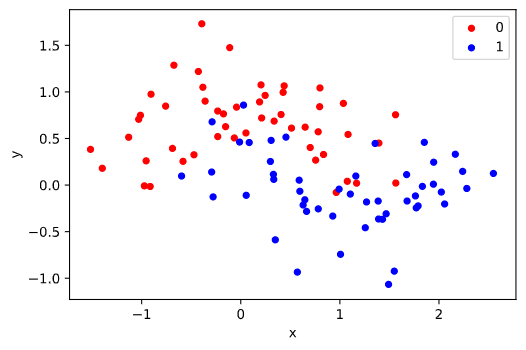}
  \caption{Moons Noise 0.25}
  \label{fig:moons25}
\end{subfigure}
\begin{subfigure}{0.4\textwidth}
  \includegraphics[width=\linewidth]{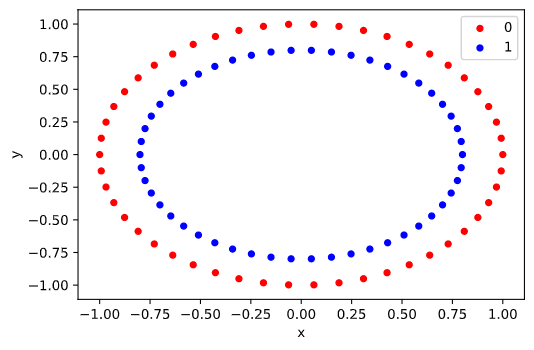}
  \caption{Circle Noise 0.0}
  \label{fig:circle0}
\end{subfigure}%
\begin{subfigure}{0.4\textwidth}
  \includegraphics[width=\linewidth]{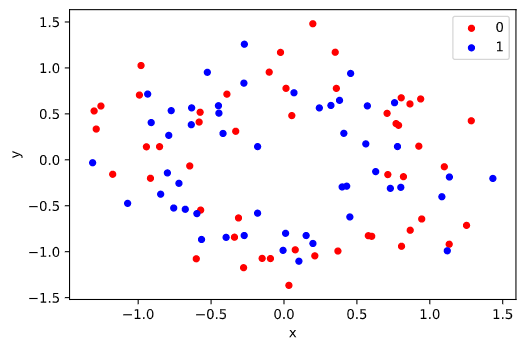}
  \caption{Circle Noise 0.25}
  \label{fig:circle25}
\end{subfigure}
\caption{Synthetic Datasets for Testing}
\label{fig:toyDatasets}
\end{figure}

The toy datasets are Moons and Circle with 0.0 and 0.25 of Gaussian standard deviation noise added to the data \ref{fig:toyDatasets}. The NBHL-BC parameter values are shown in table \ref{tab:parametersToyDataset}, and the classification accuracy results are shown in table
\ref{tab:resultsToyDataset}. These datasets were used because present clear data patterns where traditional algorithms reduce their effectiveness. In the case of Decision tree, we use gini index as quality measure without pruning method. In the case of Random Forest, we use gini index as split criterion and 100 trees. In the case of MLP, we use 2 hidden layer with 10 nodes and 100 interactions for dataset without noise and 500 interactions with noise.

\begin{table}[h!]
    \centering
    \begin{tabular}{ |c|c|c|c|c|c|  }
     \hline
     Dataset & $k$ & $e$ & $b$ & $\alpha$ & accuracy\\
     \hline
     Moons 0.00  &    5   & 0.5 & 5 & 1.0 &\bftab 100.0\\
     Moons 0.25  &    8  & 0.0 & 10 & 1.0 &\bftab 97.0\\
     Circle 0.00 &    1   & 0.5 & 1 & 1.0 &\bftab 100.0\\
     Circle 0.25 &    5   & 0.5 & 1 & 1.0 & 64.0\\
     \hline
     Moons 0.00  &    5   & 0.5 & 5 & 0.5&\bftab 100.0 \\
     Moons 0.25  &    9  & 0.0 & 10 & 0.5& 96.0 \\
     Circle 0.00 &    1   & 0.0 & 1 & 0.5&\bftab 100.0 \\
     Circle 0.25 &    5   & 0.5 & 1 & 0.5 &\bftab 65.0\\
     \hline
     Moons 0.00  &    5   & 0.5 & 5 & 0.0&\bftab 100.0 \\
     Moons 0.25  &    9  & 0.0 & 10 & 0.0 & 96.0\\
     Circle 0.00 &    1   & 0.0 & 1 & 0.0 &\bftab 100.0\\
     Circle 0.25 &    5   & 0.5 & 1 & 0.0 & 64.0\\
     \hline
    \end{tabular}
    \caption{Parameter values used by our algorithm (NBHL-BC) in Toy Datasets}
    \label{tab:parametersToyDataset}
\end{table}

We use in the first group $\alpha = 1.0$ because we want to evaluate just the structural methodology using $betweenness\ centrality$.
In the second group, we combine both strategies with $\alpha = 0.5$ and we got a small improvement on the dataset Circle 0.25.
In the last group, we use just the number of links and we got similar results but there is a reduction of the accuracy in Moons 0.25.
In some cases, we need to remove the property of $\epsilon-radius$ using $e=0.0$ and increase the quantity of $k$ neighbors in Moons with 0.25 noise. The $b$ similar $betweenness\ centrality$ nodes were kept in all the tests because other values reduce accuracy.

\begin{table}[h!]
    \centering
    \begin{tabular}{ |c|c|c|c|c|  }
     \hline
     &              MLP  &   DT    &    RF&       NBHL-BC\\
     \hline
     Moons 0.00  &  94.0&    95.0  &   98.0 &\bftab       100.0 \\
     Moons 0.25  &  84.0&    85.0  &   91.0 &\bftab        97.0  \\
     Circle 0.00 &  90.0&   92.0 &   91.0 &\bftab        100.0 \\
     Circle 0.25 &  62.0&    56.0  &   56.0 &\bftab       65.0  \\
     \hline
    \end{tabular}
    \caption{Classification accuracy of the NBHL-BC compared to Multi Layer Perceptron (MLP), Decision Tree (DT), and Random Forest (RF) in Toy Datasets}
    \label{tab:resultsToyDataset}
\end{table}

In this simulations, our algorithm presents the best results in all the datasets. Specially in the the circle dataset with noise 0.25, which is the most difficult case, our algorithm presents better classification accuracy than other techniques under comparison. 

\section{Experimental Results on Real Datasets}

In this section, we are going to present the results of the NBHL-BC technique on UCI classification datasets \cite{Dua:2019} . Also, we will compare our results with other algorithms . We tested our algorithm against Multi Layer Perceptron (MLP) \cite{RPROP-MLP}, Decision Tree C4.5 (DT) \cite{DecisionTree}, Random Forest (RF) \cite{breiman2001random}, and the Network Base High Level Technique (NBHL) \cite{tiago2018hldc}. 

The algorithms are tested splitting each dataset in two sub data sets, for training and testing with a proportion of 75\% and 25\% respectively following an stratified sampling using python as programming language and Scikit-learn library for algorithms  .

The datasets used are shown in table \ref{tab:datasets} with the number of instances, attributes and classes. These datasets were selected because the previous high-level algorithm used them. The NBHL-BC parameter values are given in table \ref{tab:parametersDataset}, and classification accuracy results are presented in table
\ref{tab:results}.

\begin{table}[h!]
    \centering
    \begin{tabular}{ |c|c|c|c|  }
     \hline
     Dataset & Instances  & Attributes & Classes\\
     \hline
     
     Glass &  214 & 9 & 6 \\
     Iris &  150 & 4 & 3 \\
     Pima &  768 & 8 & 2 \\\
     Teaching &  151 & 5 & 3 \\
     Wine &  178 & 13 & 3 \\
     Yeast &  1484 & 8 & 10 \\
     Zoo &  101 & 16 & 7 \\
     
     \hline
    \end{tabular}
    \caption{Information about the UCI classification dataset used on these project}
    \label{tab:datasets}
\end{table}

\begin{table}[h!]
    \centering
    \begin{tabular}{ |c|c|c|c|c|  }
     \hline
     Dataset & $k$ & $e$ & $b$ & $\alpha$ \\
     \hline
     
     Glass &    1   & 0.0 & 1 & 1.0 \\
     Iris  &    7  & 0.0 & 3 & 1.0 \\
     Pima  &    8   & 0.0 & 4 & 1.0 \\
     Teaching&  5   & 0.0 & 5 & 1.0 \\
     Wine &     12  & 0.0 & 5 & 1.0 \\
     Yeast &    14   & 0.0 & 3 & 0.5 \\
     Zoo &      1   & 0.0 & 1 & 1.0 \\
     
     \hline
    \end{tabular}
    \caption{Parameter values used by our algorithm (NBHL-BC) in UCI datasets}
    \label{tab:parametersDataset}
\end{table}

\begin{table}[h!]
    \centering
    \begin{tabular}{ |c|c|c|c|c|c|  }
     \hline
     &MLP&DT&RF&NBHL&NBHL-BC\\
     \hline
     Glass      &  69.231 & 63.077  &\bftab  75.385 & 66.700 & 69.231 \\
     Iris       &  93.333 & 93.333  &  93.333 &\bftab 97.400 & 95.556 \\
     Pima       &  74.892 & 69.264  &\bftab  77.056 & 73.400 &\bftab 77.056 \\
     Teaching   &  52.174 & 52.174  &  60.870  & 55.300 &\bftab 65.217 \\
     Wine       &  96.296 & 92.593  &\bftab  98.148 & 80.000 &\bftab 98.148 \\
     Yeast      &  59.641 & 48.430  &\bftab  61.883 & 36.700 & 54.036 \\
     Zoo        &  96.774 & 96.774  &  96.774 &\bftab 100.00&\bftab 100.00 \\
     
     \hline
    \end{tabular}
    \caption{Classification accuracy results of the NBHL-BC compared to Multi Layer Perceptron (MLP), Decision Tree C4.5 (DT), Random Forest (RF), and Network Base High Level Classification (NBHL) using the testing dataset.}
    \label{tab:results}
\end{table}

Our algorithm presents a good performance in all the datasets compared to other algorithms. In four cases, our algorithm presents the best results. Just in case of Iris dataset, another high level classification algorithm NBHL is better than the proposed one.

Moreover, the $\alpha$ parameter that regulates the weight between the $betweenness$ measure and number of links in 6 of the 7 datasets is 1.0 that means that the algorithm just use the $betweenness$. In the dataset Yeast, it was required an $\alpha=0.5$ that means that give same importance between $betweenness$ and number of links.
In table \ref{tab:alphaVariations}, we tested UCI Wine dataset \cite{Dua:2019} using 10-fold cross validation with different values for $\alpha$. The accuracy with only links number $\alpha=0.0$ is quite lower than $\alpha=1.0$, and the best result is mixing both techniques with $\alpha=0.4$. The $b$ parameter that evaluates the number of nodes with the lower $betweenness\ centrality$ difference with respect to the inserted node were kept constant.

\begin{table}[h!]
    \centering
    \begin{tabular}{ |c|c|c|c|c|c|  }
     \hline
     Dataset & $k$ & $e$ & $b$ & $\alpha$ & accuracy\\
     \hline
     Wine  &    8   & 0.5 & 5 & 0.0 & 95.492\\
     Wine  &    8   & 0.5 & 5 & 0.1 & 96.619\\
     Wine  &    8   & 0.5 & 5 & 0.2 & 96.619\\
     Wine  &    8   & 0.5 & 5 & 0.3 & 96.619\\
     Wine  &    8   & 0.5 & 5 & 0.4 &\bftab 97.175\\
     Wine  &    8   & 0.5 & 5 & 0.5 & 96.619\\
     Wine  &    8   & 0.5 & 5 & 0.6 & 96.063\\
     Wine  &    8   & 0.5 & 5 & 0.7 & 96.048\\
     Wine  &    8   & 0.5 & 5 & 0.8 & 95.508\\
     Wine  &    8   & 0.5 & 5 & 0.9 & 96.619\\
     Wine  &    8   & 0.5 & 5 & 1.0 & 96.048\\
     \hline
    \end{tabular}
    \caption{ Results of 10-folds cross validation in UCI Wine dataset with the training dataset. }
    \label{tab:alphaVariations}
\end{table}

\section {Conclusions}

 In this paper, we describe a new technique for high-level classification using $betweenness\ centrality$ property. We propose that nodes with similar $betweenness\ centrality$ could determinate the new untagged instance belongs to. This measure provides the importance of each node in the sub-graph communication. We exploit this property to classify a new node into a sub-graph that presents a similar communication structure. We test this algorithm in 4 toy datasets and 7 real datasets and the results are promising.
 
  As further works, we think that it is needed some procedures to reduce the noisy instances, and attributes that could produce disconnected sub graphs. Also, it is needed a way to detect the best parameters for $k,p,e,$ and $\alpha$ perhaps following an optimization approach like particle swarm.
\bibliographystyle{sbc}
\bibliography{sbc-template}

\end{document}